\documentclass[letterpaper, 10 pt, conference]{ieeeconf}  

\IEEEoverridecommandlockouts                              

\usepackage{graphics} 
\usepackage{epsfig} 
\usepackage{mathptmx} 
\usepackage{times} 
\usepackage{amsmath} 
\usepackage{amssymb}  
\usepackage{xcolor}  
\usepackage{soul}   
\usepackage[pagebackref=true,breaklinks=true,letterpaper=true,colorlinks,bookmarks=false]{hyperref}

\usepackage{multirow}
\usepackage{multicol}
\usepackage{mathtools}
\usepackage{graphicx}
\usepackage{caption}
\usepackage{array}

\usepackage{cite}

\usepackage{bbding}

\title{\LARGE \bf
PEGG-Net: Pixel-Wise Efficient Grasp Generation in Complex Scenes
}

\author{Haozhe Wang$^{1,2*}$, Zhiyang Liu$^{1*}$, Lei Zhou$^{1\dagger}$, Huan Yin$^{3}$, and Marcelo H. Ang Jr.$^{1}$
\thanks{*The authors contributed equally to this work.}
\thanks{$\dagger$Corresponding Author.}
\thanks{$^{1}$Haozhe Wang is with the Integrative Sciences and Engineering Programme, National University of Singapore Graduate School, 119077, Singapore.}%
\thanks{$^{2}$Authors are with the Department of Mechanical Engineering, National University of Singapore, 117608, Singapore. {\tt\small \{wang\_haozhe, zhiyang, leizhou\}@u.nus.edu}, {\tt\small mpeang@nus.edu.sg}}%
\thanks{$^3$Huan Yin is with the Department of Electronic and Computer Engineering, Hong Kong University of Science and Technology, Hong Kong. {\tt\small eehyin@ust.hk} }
}

\begin{document}

\maketitle
\thispagestyle{empty}
\pagestyle{empty}





\begin{abstract}
    
Vision-based grasp estimation is an essential part of robotic manipulation tasks in the real world. Existing planar grasp estimation algorithms have been demonstrated to work well in relatively simple scenes. But when it comes to complex scenes, such as cluttered scenes with messy backgrounds and moving objects, the algorithms from previous works are prone to generate inaccurate and unstable grasping contact points. In this work, we first study the existing planar grasp estimation algorithms and analyze the related challenges in complex scenes. Secondly, we design a Pixel-wise Efficient Grasp Generation Network (PEGG-Net) to tackle the problem of grasping in complex scenes. PEGG-Net can achieve improved state-of-the-art performance on the Cornell dataset (98.9\%) and second-best performance on the Jacquard dataset (93.8\%), outperforming other existing algorithms without the introduction of complex structures. Thirdly, PEGG-Net could operate in a closed-loop manner for added robustness in dynamic environments using position-based visual servoing (PBVS). Finally, we conduct real-world experiments on static, dynamic, and cluttered objects in different complex scenes. The results show that our proposed network achieves a high success rate in grasping irregular objects, household objects, and workshop tools. To benefit the community, our trained model and supplementary materials are available at \url{https://github.com/HZWang96/PEGG-Net}.

\end{abstract}


\section{Introduction}

Grasp generation in cluttered and dynamic environments is a challenging problem in robotics. Vision-based robotic grasping uses inputs from depth cameras, RGB-D cameras, or other similar sensors to generate a reliable grasping approach for manipulators to grab and lift objects. Existing grasp detection techniques can be divided into two main categories: analytical methods and data-driven methods \cite{Du2021survey,Li2019survey,Duan2021survey,Newbury2022survey}. An analytical method uses kinematic, kinetic, and mechanical analysis of the geometry and pose of the object as a hard-coded strategy to ascertain the grasping position. The data-driven approach is primarily based on machine learning techniques, which learn from large quantities of grasping experience data to determine the best grasping position of objects. Since there are bound to be many unknown objects in practical robot applications, the data-driven approach would be preferable to the analytical method for grasping unknown objects. With the proliferation of deep learning, many research works~\cite{doi:10.1177/0278364919868017,morrison2018closing,kumra2020antipodal} have proposed learning-based methods to achieve object-agnostic grasping.
 
Most previous studies~\cite{morrison2018closing,kumra2020antipodal,chu2018real,article,Morrison2020LearningRR,redmon2015realtime} designed networks and demonstrated them under constraint conditions that are too ideal for many real-world scenarios. However, real-world environments are often more cluttered and may contain dynamically moving objects, which can limit the practical usefulness of existing models in applications such as manufacturing, warehousing, and domestic robotics. To address the issue of grasping in complex scenarios, this paper proposes a new robust and efficient approach to address these challenges and improve the success rate of grasping objects in real-world scenarios.


\begin{figure}[t]
	\centering
	\includegraphics[scale=0.45]{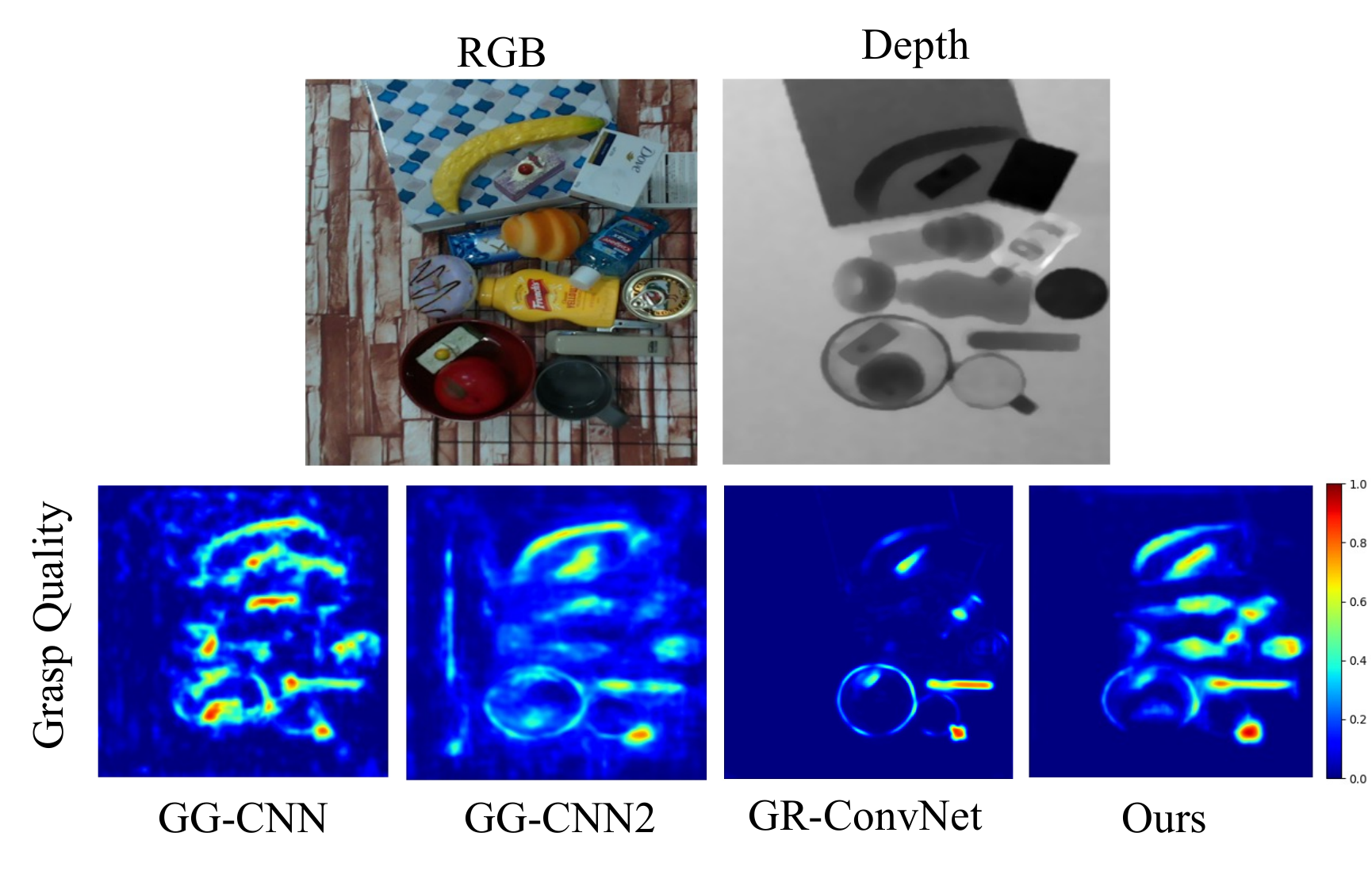}
	\caption{Visualization of predictions by GG-CNN~\cite{morrison2018closing}, GG-CNN2~\cite{Morrison2020LearningRR}, GRConvNet~\cite{kumra2020antipodal}, and PEGG-Net. PEGG-Net could generate higher-quality grasp poses compared to previous works. This demonstrates the robustness of our network in complex scenes.}
	\label{teaser}
 \vspace{-4mm}
\end{figure}



In this paper, we design a lightweight network that can predict the 2D planar grasp of a novel object in a complex scene without prior knowledge of the pose of the object itself. More concretely, a complex scene could be a scene with a random background and static objects in clutter or dynamic objects on top of it. Our proposed network could generate grasping poses effectively and efficiently based on its understanding of complex scenes. Overall, the two main contributions of this paper can be summarized as follows:
\begin{enumerate}
    \item We propose PEGG-Net, a robust and efficient pixel-wise grasp generation convolutional neural network (CNN), which is capable of supporting robotic grasping in complex scenes using either RGB or depth-aided images. PEGG-Net achieves state-of-the-art and second-best accuracy of 98.9\% and 93.8\% on the public Cornell and Jacquard grasping datasets, respectively.
    \item {With the grasp poses generated by PEGG-Net, we design and implement a network-integrated grasping system}. This makes PEGG-Net more applicable in complex real-world scenes. {We conduct real-world grasping experiments in static, dynamic, and cluttered, complex scenes using irregular objects, household objects, and workshop tools and achieve an overall grasp success rate of 90.2\%.}
\end{enumerate}

We demonstrate the robustness and generalization capabilities of PEGG-Net and the PEGG-Net-aided control system in complex scenes using Kinova Movo with a RealSense L515 camera. We test our grasping system using objects with adversarial geometry from the DexNet 2.0 dataset~\cite{mahler2017dexnet}, household objects, and workshop tools. All objects used in our real-world test are unseen during the training stage. We run our experiments under static, dynamic, and cluttered conditions and achieve higher levels of accuracy compared to previous works.
    
{Overall, PEGG-Net, together with our network-integrated control system, achieves general improvements in grasp stability, accuracy, efficiency, and generalization capability compared to previous works in this area.} Our work enhances the ability of the grasp generation model to handle complex real-world scenes without introducing new limitations.

\section{Related Works}

\subsection{{Deep Learning Methods for Planar Grasping of Novel Objects}}
Deep learning has been instrumental in advancing the field of computer vision in general and hence vision-based robotic grasping as well. Many recent works applying deep learning techniques to grasp novel objects have achieved great success. 

Regression-based methods proposed by Redmon and Angelova~\cite{redmon2015realtime} and Kumra \textit{et al.}~\cite{kumra2017robotic} used a CNN to regress a single best grasp pose from an input image with only one object in it. However, such networks could not handle cluttered environments and may generate an invalid grasp that is the average of the possible grasps for an object.
Chu \textit{et al.}~\cite{chu2018real} proposed to predict multiple grasps for multiple objects simultaneously. However, the network is too computationally intensive.

{It is important to note that most grasp prediction networks proposed in recent works have a large number of parameters (in the order of tens of millions to even hundreds of millions of parameters)~\cite{chu2018real,mahler2017dexnet,lenz2014deep,pinto2015supersizing,johns2016deep,GKNet}, which makes them very computationally intensive. Therefore most networks proposed in previous works can only be executed in an open-loop manner on resource-constrained platforms. The critical disadvantage of open-loop grasping systems is that they cannot grasp objects in dynamic environments, for example, objects moving on a conveyor belt. Even if an object is static, sensor and actuation errors can also affect the grasping accuracy of open-loop grasping systems~\cite{lenz2014deep}. Hence, we argue that adding a closed-loop grasping system is a more robust solution for many practical applications.}

{Some deep learning techniques~\cite{doi:10.1177/0278364919868017,mahler2017dexnet,lenz2014deep,5980145,pinto2015supersizing,yan2019dataefficient} first classified grasp candidates sampled from an image or point cloud, then ranked them individually to determine the best grasp pose.} Other approaches have attempted to simplify the problem by evaluating their models on scenes with either clean backgrounds~\cite{morrison2018closing,kumra2020antipodal,wu2021dense,tian2022dense} or without clutter~\cite{tian2022dense,wang2019efficient}. Doing so made it easier for their networks to predict accurate and stable grasp poses. However, in the real world, a robot's complex working environment will negatively influence the decisions made by the grasp generation network and drastically reduce the grasping confidence and the gripper orientation and width predictions from the network. Our empirical observations from reproducing previous works by Morrison \textit{et al.}~\cite{morrison2018closing}\cite{Morrison2020LearningRR} and Kumra \textit{et al.}~\cite{kumra2020antipodal} show that their networks are unable to ignore the influence of the background and will sometimes predict invalid grasping points that are in the background. We provide qualitative comparisons between our work and those by Morrison \textit{et al.} and Kumra \textit{et al.} in Fig. \ref{teaser} when performing inference on a cluttered scene with a messy and textured background. 


\subsection{Visual Servoing for Closed-Loop Grasping}
Visual servoing is a technique that uses visual feedback extracted from a vision sensor to control the motion of a robot. Visual servoing can be incorporated into a robotic grasping system to create a closed-loop control system. Unlike open-loop grasping, closed-loop grasping allows for grasp poses that can be continuously refined as the gripper approaches an object. A closed-loop robotic grasping system is much more robust against dynamic scenes or sensor and actuation errors. To achieve fast and accurate closed-loop grasping, the grasp prediction network must be able to execute in real-time (at least as fast as the sampling rate of the camera), and there should be minimal delay in the control system so that the robot can respond immediately to the changing environment.

Several previous works~\cite{morrison2018closing,lenz2014deep,sergey2016learning,viereck2017learning} have integrated visual servoing techniques into their robotic grasping systems. CNN-based controllers proposed in recent years~\cite{sergey2016learning,viereck2017learning} combine deep learning with closed-loop grasping. However, the controllers used in those works could not run faster than 5Hz, which is too slow to support closed-loop grasping in dynamic environments. On the other hand, we note that PBVS has achieved closed-loop grasping up to 50Hz~\cite{morrison2018closing}, much faster than CNN-based methods. Inspired by these studies, we combine PBVS with PEGG-Net in our closed-loop robotic grasping system.

\section{Problem Formulation}

For PEGG-Net to predict a 2D planar grasp on an arbitrary object, we adopt a grasp definition similar to other related works~\cite{morrison2018closing,kumra2020antipodal}. Given an RGB-D camera, intrinsic parameters and transformations (extrinsic parameters) between the camera and the end-effector are known. PEGG-Net aims at generating a grasp in the image coordinate frame, which is defined as:
\begin{equation}
\label{eqn:g_img}
{\hat{g}}=(\hat{\mathbf{p}},\hat{\phi},\hat{\omega},q)
\end{equation}
where $\hat{\mathbf{p}}=(u,v)$ are the 2D coordinates of the grasping point in the image frame, $\hat{\phi}$ is the rotation angle around the vertical orientation of the image, $\hat{\omega}$ is the grasping width in pixels and $q$ is the grasp quality (the probability of a successful grasp). It is worth mentioning here that our grasping width is not fixed, which is unlike other works~\cite{johns2016deep, viereck2017learning}.

To execute the grasp predicted by PEGG-Net, the grasp $\hat{g}$ in the image coordinate frame must be converted to the corresponding grasp $g$ in the world coordinate frame, which is defined as 
\begin{equation}
\label{eqn:g_real}
g=(\mathbf{p},\phi,\omega,q)
\end{equation}
where $\mathbf{p} = (x,y,z)$ is in world coordinates ($(x, y)$ is the centre of the gripper and $z$ is the depth of the grasp, which is measured using a depth camera), $\phi$ is the gripper's rotation angle about the z-axis, $\omega$ is the gripper's opening width $\omega$, and $q$ is the grasp quality. $\hat{g}$ and $g$ are governed by the following transformations:
\begin{equation}
\label{eqn:transform}
g=T_{RC}(T_{CI}(\hat{g}))
\end{equation}
where $T_{CI}$ is the transformation from the image coordinate frame to the camera coordinate frame and $T_{RC}$ is the transformation from the camera coordinate frame to the world coordinate frame. The grasp pose can be computed by using (\ref{eqn:transform}).

In this study, PEGG-Net predicts a grasp for every pixel in an image. Therefore, we can apply (\ref{eqn:g_img}), (\ref{eqn:g_real}), and (\ref{eqn:transform}) to an image with multiple grasps. The set of all possible grasps in the image coordinate frame can be denoted as:
\begin{equation}
\label{eqn:grasp_set}
\mathbf{\hat{G}} = (\hat{\Phi}, \hat{\Omega}, \mathbf{Q}) \in \mathbb{R}^{3 \times H \times W}
\end{equation}
where 
\begin{equation}
\label{eqn:non_neg_real_num}
\mathbb R_{\ge 0} = \{x \in \mathbb{R} : x \ge{0}\}
\end{equation}
\begin{equation}
\label{eqn:Phi_hat}
\hat{\Phi} = \{\hat{\phi} \in \mathbb R_{\ge 0}\}, \hat{\Phi} \in \mathbb R_{\ge 0}^{H \times W}
\end{equation}
\begin{equation}
\label{eqn:Omega_hat}
\hat{\Omega} = \{\hat{\omega} \in \mathbb R_{\ge 0}\}, \hat{\Omega} \in \mathbb R_{\ge 0}^{H \times W}
\end{equation}
\begin{equation}
\label{eqn:Q}
\mathbf{Q} = \{q \in \mathbb R_{\ge 0}\}, \mathbf{Q} \in \mathbb R_{\ge 0}^{H \times W}
\end{equation}
and $\hat{\Phi}$, $\hat{\Omega}$ and $\mathbf{Q}$ are the feature embeddings ($H \times W$) for $\hat{\phi}$, $\hat{\omega}$ and $q$, respectively.

This is a more computationally efficient way of formulating the grasping problem, as there is no need to sample and rank grasp candidates~\cite{doi:10.1177/0278364919868017,mahler2017dexnet,doi:10.1177/1687814016668077,lenz2014deep,pinto2015supersizing,yan2019dataefficient}. We can simply obtain the best grasp in the image coordinate frame by computing $\hat{g}_{best} = \underset{\mathbf{Q}}{\max}$ $\mathbf{\hat{G}}$ and obtain the best grasp in the world coordinate frame $g_{best}$ using (\ref{eqn:transform}).



\begin{figure}[t]
	\centering
	\includegraphics[scale=0.24]{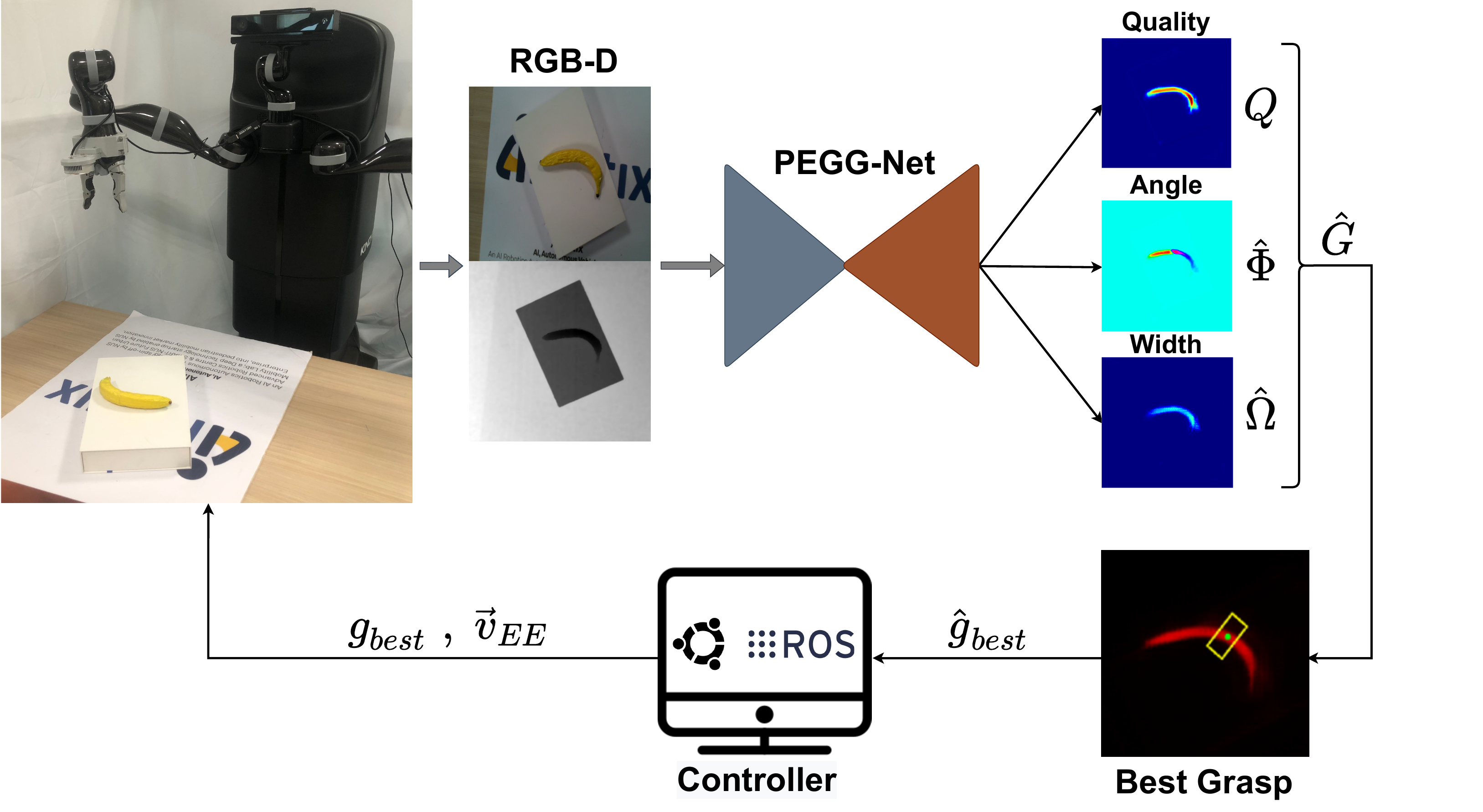}
	\caption{Overview of our closed-loop grasping pipeline using the proposed grasp generation network, PEGG-Net.}
	\label{closed-loop}
 \vspace{-4mm}
\end{figure}

\section{{Network-Integrated Grasping System}}

{We divide the entire system into two parts to better illustrate our proposed grasping system: grasp pose generation using PEGG-Net and the robot control system}. Note that these two parts are coupled together to perform real-time grasping. An overview of the  pipeline is shown in Fig.~\ref{closed-loop}.

\subsection{Network Architecture}
\begin{figure*}[t]
	\centering
	\includegraphics[width=15.96cm]{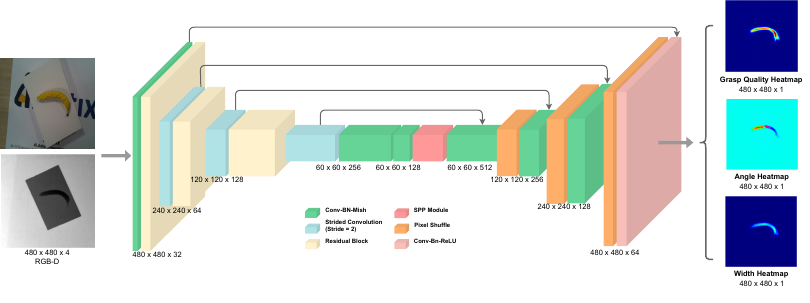}
	\caption{Architecture of PEGG-Net. PEGG-Net consists of six types of basic network blocks: Conv-BN-Mish (green), Strided Convolution (blue), Residual Block (beige), SPP Module (red), Pixel Shuffle (orange) and Conv-Bn-ReLU (pink). PEGG-Net takes RGB and depth images as inputs, and it outputs heatmaps that include grasping quality, angle, and width. This figure is better viewed in color.}
	\label{framework}
 \vspace{-4mm}
\end{figure*}
PEGG-Net aims at generating high-quality grasps in real time for complex scenes. The network is desired to be accurate and lightweight to guarantee the effectiveness and efficiency of grasping, bringing challenges to the network design in this article.

We propose to design a fully convolutional encoder-decoder network. Fig. \ref{framework} describes the network architecture of PEGG-Net. Specifically, the encoder uses residual blocks and strided convolutions to extract key features of graspable objects in the workspace. Springenberg \textit{et al.}~\cite{SpringenbergDBR14} found that replacing max-pooling layers with a convolution layer of stride 2 improves the model's accuracy. The results from our ablation study (summarized in Table \ref{abla_network}) agree with Springenberg's findings. The output from the encoder network is passed through a modified spatial pyramid pooling (SPP) module proposed by Huang and Wang~\cite{huang2019dcsppyolo}. {The SPP module can extract features at multiple scales to capture the salient parts of an image---the appearance of the graspable objects---and the background context at different levels of detail. This multi-scale representation can help the model distinguish a graspable object from the background by capturing its distinctive features at different levels of detail. This multi-scale feature representation can also help the model to generate accurate grasp poses for objects of different sizes and aspect ratios.} We use max-pooling kernels of size $5\times5$, $9\times9$, and $13\times13$. The decoder, implemented using the sub-pixel convolution method~\cite{shi2016realtime}, reconstructs the object features extracted by the encoder. Skip connections are added between the encoder and decoder layers to allow the decoder to use features extracted at various stages in the encoder to reconstruct the object features. Guided by the results from our ablation study in Table \ref{abla_network}, we use the Mish~\cite{misra2020mish} activation function in all convolution blocks except the last one, which uses the ReLU activation function, and add batch normalization to all layers. The Mish function is a smooth curve, and smooth activation functions allow for better information propagation deeper into the neural network, and thus better accuracy and generalization. Based on our results in Table \ref{abla_network}, Smooth L1 loss is used as the loss function for training our network, formulated as follows:
\begin{equation}
\label{eqn:smoothL1}
\mathcal{L}(x,y) = \frac{1}{n}\sum_{u=1}^W\sum_{v=1}^H{\ell_{u,v}}
\end{equation}
where $\ell_u,v$ is given by:
\begin{equation}
\label{eqn:smoothL1_desc}
\ell_uv = \begin{cases}
            0.5\left(x_{u,v}-y_{u,v}\right)^2 / \beta, & \text {if}\left|x_{u,v}-y_{u,v}\right|<\beta \\ 
            \left|x_{u,v}-y_{u,v}\right|-0.5 * \beta, & \text {otherwise}
         \end{cases}
\end{equation}
where $x_{u,v}$ is the prediction from PEGG-Net and $y_{u,v}$ is the ground truth label. We set $\beta=1$.
The total loss is the summation of the Smooth L1 losses computed for the grasping quality, grasping angle, and gripper width:
\begin{equation}
\label{eqn:total_losss}
\mathcal{L}(\Tilde{q}, \Tilde{\phi}, \Tilde{\omega}) = \mathcal{L}(q,\Tilde{q}) + \mathcal{L}(\phi,\Tilde{\phi}) + \mathcal{L}(\omega,\Tilde{\omega})
\end{equation}
where $\Tilde{q}$, $\Tilde{\phi}$, $\Tilde{\omega}$ are the ground truth labels for grasp quality, grasp angle, and gripper width, respectively.

The network outputs a grasp quality heatmap, a grasp angle map, and a gripper width map. The location of the pixel with the highest grasp quality (the highest probability of a successful grasp) is selected from the grasp quality heatmap, and the corresponding gripper width and the angle at that same location are used to set the grasp pose of the gripper.

Our final network has 1.38 million parameters. From Table \ref{compare_params}, we can observe that, except for GG-CNN2, our network is much smaller than those proposed in previous works. This makes PEGG-Net highly suitable for performing real-time closed-loop robotic grasping. The superiority of PEGG-Net will be validated in the following experimental section, including both effectiveness and efficiency for grasping.



\begin{table}[t]
	\captionsetup{justification=centering}
	\renewcommand\arraystretch{1.0}
	\begin{center}
		\caption{Ablation study on the combination of downsampling, activation function, and loss function to be used in PEGG-Net.}
		\label{abla_network}
		\begin{tabular}{p{4.8cm}<{\centering}|p{0.8cm}<{\centering}p{1.3cm}<{\centering}}
			\hline
			\hline
			\multirow{2}*{Combination} & \multicolumn{2}{c}{Accuracy} \\
		    & D (\%) & RGB-D(\%) \\
			\hline
			Max-Pool + Mish + Smooth L1 loss & 51.7 & 78.7 \\
                \hline
                Strided Conv. + ReLU + Smooth L1 loss & 71.9 & 71.9 \\
			\hline
			Strided Conv. + Mish + MSE loss & 62.9 & 83.1 \\
			 \hline
			Strided Conv. + Mish + Smooth L1 loss & \textbf{87.6} & \textbf{91.0} \\
			\hline
			\hline
		\end{tabular}
	\end{center}
\vspace{-4mm}
\end{table}

\begin{table}[t]
	\captionsetup{justification=centering}
	\renewcommand\arraystretch{1.0}
	\begin{center}
		\caption{Network size comparison of grasping methods.}
		\label{compare_params}
		\begin{tabular}{p{1.5cm}<{\centering}|p{2.6cm}<{\centering}|p{2.7cm}<{\centering}}
			\hline
			\hline
			Author & Method & Number of Parameters (Approx.) \\
			\hline
			Morrison~\cite{Morrison2020LearningRR} & GG-CNN2 & \textbf{66 k} \\
			\hline
			Kumra~\cite{kumra2020antipodal} & GR-ConvNet & 1.9 million \\
	        \hline
			Tian~\cite{tian2022dense} & RGB-D Dense Fusion & 7.24 million \\
			\hline
            {Xu ~\cite{GKNet}} & {GKNet} & {15.74 million} \\
            \hline
			Mahler~\cite{mahler2017dexnet} & DexNet 2.0 & 16 million \\
			\hline
			Pinto~\cite{pinto2015supersizing} & AlexNet & 61 million \\
			\hline
			Chu~\cite{chu2018real} & ResNet-50 + GPN & 216 million \\
			\hline
			Ours & PEGG-Net & \underline{1.38 million} \\
			\hline
			\hline
		\end{tabular}
	\end{center}
\vspace{-6mm}
\end{table}

\subsection{{Implementation of the Robot Control System}}

We implement a PBVS controller~\cite{VS_survey} to perform closed-loop grasping. We move the end-effector to the final grasping position simply by controlling its velocity. {Our network-integrated robotic grasping system can achieve closed-loop control at up to 50Hz, which could be particularly beneficial in dynamic environments where fast and reliable grasping is critical, such as in manufacturing or warehouse environments.}

Fig \ref{closed-loop} illustrates how the closed-loop control system is integrated with PEGG-Net. The RealSense camera is initially 50cm above the workspace. The RGB-D images are captured and processed by PEGG-Net in real time. There may be many grasping points on the object with similar grasp quality scores. To avoid switching rapidly between grasping points and confusing the controller, we compute 5 grasps from the highest local maxima of $\mathbf{\hat{G}}$ and select which is closest (in the image coordinate frame) to the grasp location predicted by PEGG-Net in the previous iteration. The controller publishes a 6D velocity signal $\Vec{v}_{EE}$ to the ROS topic, controlling the end-effector's motion.

At the beginning of each grasp attempt, the system is initialized to track the global maxima of $\mathbf{Q}$. PEGG-Net and the PBVS controller form a feedback loop, enabling PEGG-Net to perform closed-loop grasping in dynamic environments. The gripper fingers are controlled via velocity signals sent from the task controller. Control is stopped when the grasp pose is reached. A video is attached with this article for a better understanding of our closed-loop control system.

\section{Experiments}

The proposed system is validated with both public datasets and a real-world robotic manipulator. Specifically, public datasets are used to evaluate the grasp generation performance of PEGG-Net (Section~\ref{sec:VA} and~\ref{sec:VB}); real-world experiments are conducted to test the proposed PEGG-Net and closed-loop control system (Section~\ref{sec:VC}, \ref{sec:VD} and~\ref{sec:VE}).

\subsection{Training Strategy}
\label{sec:VA}
We train PEGG-Net on the Cornell and Jacquard datasets using an NVIDIA Tesla V100 GPU with 16GB of memory. The Cornell dataset has 885 images with 8,019 grasp annotations and the Jacquard dataset has 54,485 images with 1,181,330 unique grasp annotations. For both Cornell and Jacquard datasets, we use 90\% of the images for training and 10\% for evaluation. We also augment both images from the Cornell dataset using random crops and zooms. Our network is trained using the Adam optimizer~\cite{kingma2017adam} with a batch size of 8.

\subsection{Evaluation on Public Datasets}
\label{sec:VB}

To compare our results with previous works on datasets, we use the rectangle metric proposed by Jiang \textit{et al.}~\cite{5980145}. The rectangle metric states that a valid grasp prediction must satisfy the following two conditions:
\begin{enumerate}
  \item The intersection over union (IoU) score between the ground truth grasp rectangle and the predicted grasp rectangle is more than 25\%.
  \item The offset between the rotation angle $\phi$ of the ground truth grasp rectangle and the predicted grasp rectangle is less than 30$^{\circ}$.
\end{enumerate}

Table \ref{Cornell} shows the performance of PEGG-Net on the Cornell dataset for different modalities and input image sizes. We compare the performance of PEGG-Net with other planar grasp generation methods proposed in previous works that have not been demonstrated to perform well in complex scenes.


All input images used for evaluating our network and the networks designed in prior works have been center-cropped from the original image size of $640 \times 480$ in the Cornell dataset to the sizes listed in Table \ref{Cornell}. When evaluated on an object-wise split, our network outperforms almost all other networks for each modality and image size. PEGG-Net achieves remarkable performance even when we use images of size $480 \times 480$, which includes the background distractions present in each image in the Cornell dataset. On the other hand, the accuracy of GG-CNN \cite{morrison2018closing}, GG-CNN2 \cite{Morrison2020LearningRR}, and GR-ConvNet \cite{kumra2020antipodal} drastically decreases when trained on the Cornell dataset with a larger input size ($480 \times 480$) compared to their original and more aggressively center-cropped images sizes ($300 \times 300$ for GG-CNN and GG-CNN2 and $224 \times 224$ for GR-ConvNet). In contrast, the accuracy of PEGG-Net is consistently high for different input sizes. 


\begin{figure*}[t]
	\centering
	\includegraphics[width=\linewidth]{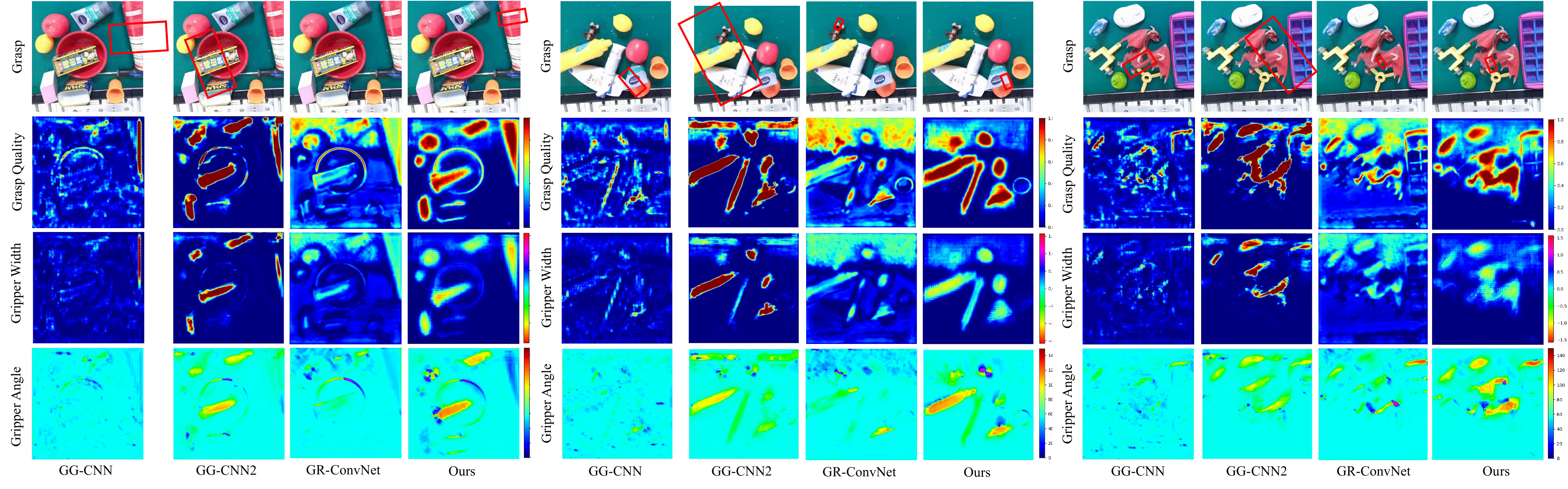}
	\caption{{A qualitative comparison of the grasp predictions of GG-CNN~\cite{morrison2018closing}, GG-CNN2~\cite{Morrison2020LearningRR}, and GR-ConvNet~\cite{kumra2020antipodal} with PEGG-Net in complex cluttered scenes. The images are from the GraspNet-1Billion benchmark dataset \cite{fang2020graspnet}. Note that all the models were not trained on the GraspNet-1Billion dataset as we want to evaluate and compare how well these models can generalize to an unseen environment with novel objects. The best grasp poses (the red rectangles) predicted by the respective models are displayed on the RGB images. PEGG-Net generates very accurate grasps despite the cluttered scene. The grasp quality of GG-CNN is much lower than PEGG-Net, the gripper angle and width predicted by GG-CNN2 and GR-ConvNet are inaccurate. This figure is best viewed zoomed in.}}
	\label{qualitative_comparison}
\vspace{-2mm}
\end{figure*}

\begin{figure}[t]
	\centering
	\includegraphics[scale=0.4]{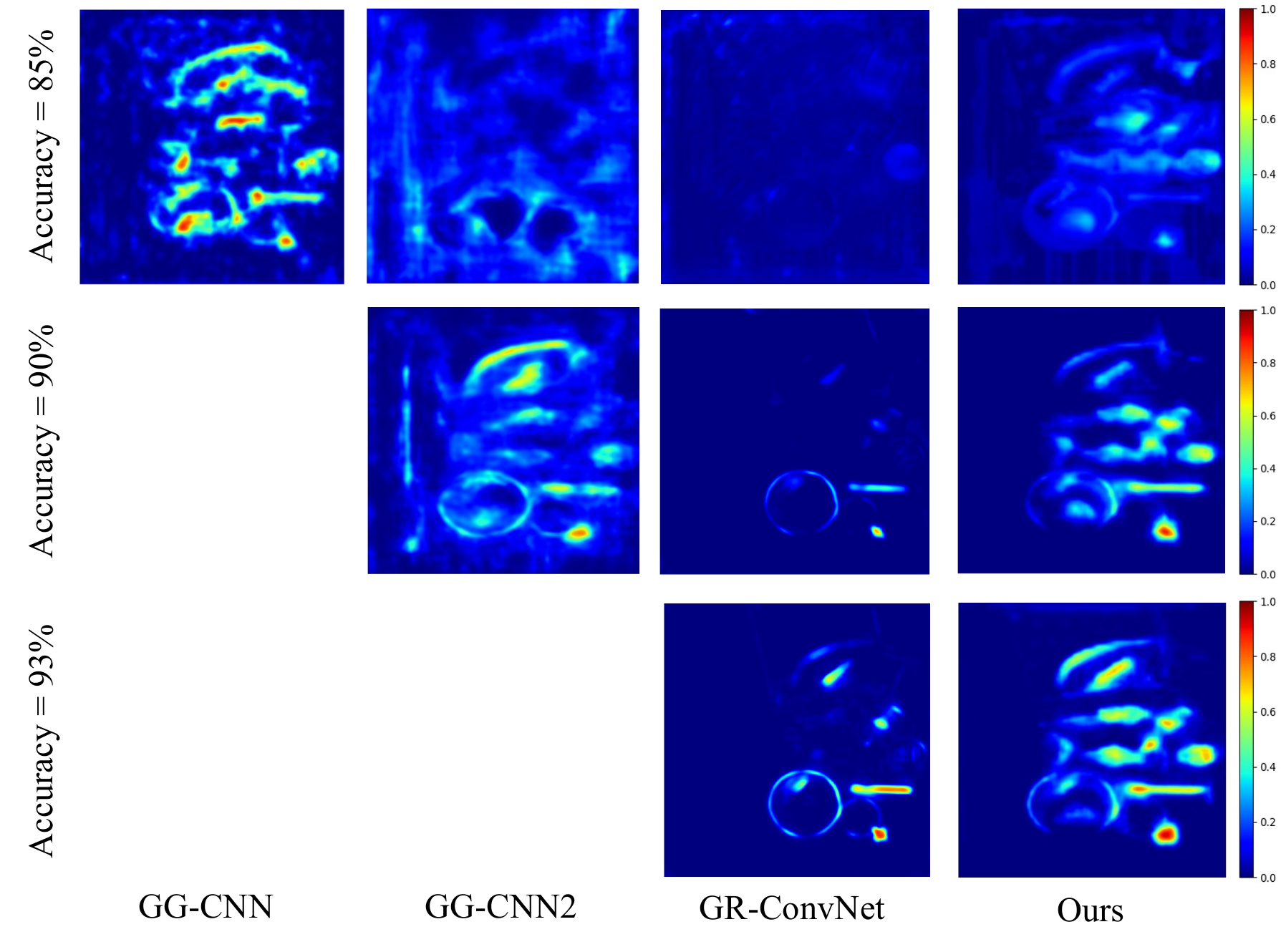}
        \caption{Grasp quality heatmaps generated by GG-CNN, GG-CNN2, GRConvNet, and PEGG-Net in a complex scene (same scene as the one in Fig. \ref{teaser}). All networks are trained and evaluated on the Jacquard dataset. The accuracy of GG-CNN and GG-CNN2 cannot increase beyond 85\% and 90\%, respectively. As the accuracy of the networks increases from 85\% to 93\%, GG-CNN and GG-CNN2 are severely impacted by background interference. Although the background interference impacts GR-ConvNet to a lesser extent, it is also less confident about how to grasp the several objects in the scene. PEGG-Net remains robust in this situation as it can predict a high grasp quality for the objects in the scene and ignores the background interference.}
	\label{clutter_comparison}
\end{figure}




Table~\ref{Jacquard} shows the performance of PEGG-Net on the Jacquard dataset for different modalities compared to previous networks proposed by Morrison \textit{et al.}~\cite{morrison2018closing, Morrison2020LearningRR} and Kumra \textit{et al.}~\cite{kumra2020antipodal}. Although PEGG-Net is slightly less accurate than GR-ConvNet and GR-ConvNetv2, this does not affect its real-world grasping performance. We illustrate using Fig. \ref{clutter_comparison} and also demonstrate through our real-world experiments in the subsequent sections that better performance on the datasets does not necessarily imply better real-world performance. Actually, achieving the best trade-off between speed and accuracy is the most crucial for closed-loop robotic grasping. PEGG-Net achieves a better trade-off between speed and accuracy than GR-ConvNet as PEGG-Net has approximately 27\% fewer parameters but is no more than 2\% less accurate than GR-ConvNet.

{In Fig. \ref{qualitative_comparison}, we provide evidence of PEGG-Net's performance in complex, cluttered scenes, using examples from the GraspNet-1Billion dataset~\cite{fang2020graspnet}. This dataset provides a more realistic and challenging environment for robotic grasping, with objects in various positions and orientations and with occlusions and clutter. These images mimic a real-world robotic grasping environment much more closely compared to the Cornell and Jacquard datasets. While PEGG-Net was not specifically trained on this dataset, we find that it is still able to accurately predict grasping poses in these challenging scenes. This indicates the robustness and generalizability of our proposed method beyond the simplicity of the training data and to a more complex real-world environment.}

\subsection{Set-up in the Real-world}
\label{sec:VC}

For our real-world experiments, we use a 7-DoF robotic arm on the Kinova Movo mobile manipulator fitted with a Kinova KG-Series 3-fingered gripper. We removed one finger and only used the remaining two fingers that are parallel to each other so that we can fairly compare PEGG-Net to other grasp generation models. The fingertips are perpendicular to the table. As shown in Fig. \ref{closed-loop}, an Intel RealSense L515 LiDAR depth camera is mounted on its wrist. We use the hand-eye calibration method to calibrate this setup before the experiment.



All computations of the grasp generation network are performed on a computer running Ubuntu 16.04 with a 3.7 GHz AMD Ryzen 5900X CPU and NVIDIA GeForce RTX 3080 GPU. On this platform, our network takes 10ms to compute a single RGB-D frame of size $480 \times 480$, and the entire grasping pipeline (including all pre-processing and post-processing computations) takes 20ms (50Hz). We use RGB-D images for all our real-world grasping experiments.

\begin{figure}[t]
	\centering
	\includegraphics[width=7cm]{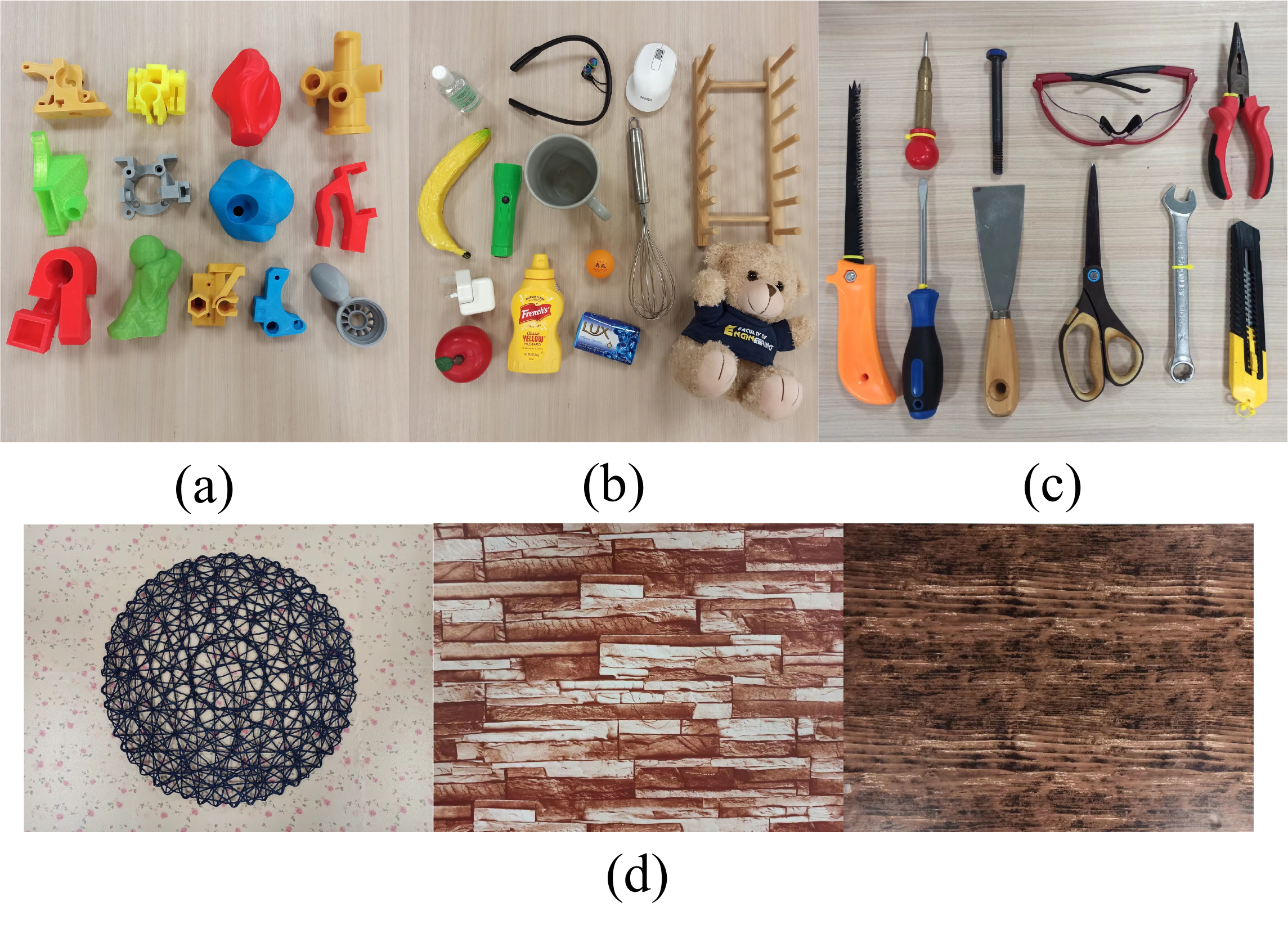}
	\caption{Objects and backgrounds used for our grasping experiments in the real world. (a) 13 adversarial objects from the DexNet 2.0 dataset~\cite{mahler2017dexnet}. (b) 14 household objects. (c) 10 workshop tools. (d) 3 backgrounds with different designs and textures.}
	\label{Objects}
\vspace{-6mm}
\end{figure}

\begin{table*}[t]
	\captionsetup{justification=centering}
	\renewcommand\arraystretch{1.0}
	\begin{center}
		\caption{Object-wise grasp detection accuracy on the Cornell dataset. * indicates the results we obtained after re-training the network using open-source code provided by the authors of prior works. The speed indicates the single image processing time of the network on the stated GPU.}
		\label{Cornell}
		\begin{tabular}{p{1.5cm}<{\centering}|p{3.5cm}<{\centering}|p{1.0cm}<{\centering}|p{2.0cm}<{\centering}|p{1.6cm}<{\centering}|p{1.5cm}<{\centering}|p{3.0cm}<{\centering}}
			\hline
			\hline
		   Author & Algorithm & Modality & Input Size & Accuracy (\%) & Speed (ms) & Hardware (GPU) \\
			 & & & & & & \\
			 \hline
              Wang~\cite{wang2019efficient} & GPWRG & D & $400 \times 400$ & 91.0 & 8 & GeForce GTX 1060\\
              \hline
              Tian~\cite{tian2022dense} & RGB-D Dense Fusion & RGB-D & $224 \times 224$ & 98.9 & 15 & GeForce RTX 3080\\
              \hline
              {Park~\cite{park2019realtime}} & {REM, DarkNet-19} & {RGB-D} & {$360 \times 360$} & {98.6} & {20} & {GeForce GTX 1080 Ti}\\
              \hline
              {Xu ~\cite{GKNet}} & {GKNet} & {RG-D} & {$256 \times 256$} & {96.9} & {17} & {GeForce RTX 3080}\\
              \hline
			 \multirow{3}*{Kumra~\cite{kumra2020antipodal}} & \multirow{3}*{GR-ConvNet} & D & \multirow{2}*{$224 \times 224$} & 94.3 & 19 & \multirow{2}*{GeForce GTX 1080 Ti} \\
			 & & RGB-D & & 96.6 & 20 & \\
              & & & & & & \\
              & & RGB-D & $480 \times 480$ & 70.8* & 9 & GeForce RTX 3080\\
			 \hline
          {Kumra~\cite{GR-ConvNetv2}} & {GR-ConvNetv2} & {RGB-D} & {$224 \times 224$} & {97.7} & {20} & 
          {GeForce GTX 1080 Ti}\\
          \hline
			 \multirow{2}*{Morrison~\cite{morrison2018closing}} & \multirow{2}*{GG-CNN} & D & \multirow{2}*{$300 \times 300$} & 76.4 & 19 & GeForce GTX 1070\\
             & & RGB-D & & \textbf{97.8*} & 3.7 & GeForce RTX 3080\\
             & & & & & & \\
             & & RGB-D & $480 \times 480$ & 74.2* & 3.7 & GeForce RTX 3080\\
			 \hline
			 \multirow{3}*{Morrison~\cite{Morrison2020LearningRR}} & \multirow{3}*{GG-CNN2} & D & \multirow{2}*{$300 \times 300$} & 71.9 & 20 & GeForce GTX 1070\\
             & & RGB-D & & \underline{95.5*} & 3.4 & GeForce RTX 3080 \\
             & & & & & & \\
             & & RGB-D & $480 \times 480$ & 88.8* & 3.4 & GeForce RTX 3080 \\
			 \hline
			 \multirow{11}*{Ours} & \multirow{11}*{PEGG-Net} & D & \multirow{2}*{$224 \times 224$} & \textbf{94.4} & 3.1 & \multirow{11}*{GeForce RTX 3080} \\ 
			 & & RGB-D & & \textbf{98.9} & 3.1 & \\
			 & & & & & & \\
             & & D & \multirow{2}*{$304 \times 304$} & 93.3 & 3.8 & \\
             & & RGB-D & & 94.4 & 3.8 & \\
             & & & & & & \\
             & & D & \multirow{2}*{$320 \times 320$} & \textbf{94.4} & 4.3 & \\
             & & RGB-D & & \textbf{96.6} & 4.3 & \\
             & & & & & & \\
             & & D & \multirow{2}*{$480 \times 480$}  & \textbf{87.6} & 10 & \\
             & & RGB-D & & \textbf{91.0} & 10 & \\
			\hline
			\hline
		\end{tabular}
	\end{center}
\vspace{-6mm}
\end{table*}

\begin{table}[t]
	\captionsetup{justification=centering}
	\renewcommand\arraystretch{1.0}
	\begin{center}
		\caption{Detection accuracy on the Jacquard dataset.}
		\label{Jacquard}
		\begin{tabular}{p{2.0cm}<{\centering}|p{1.8cm}<{\centering}|p{1.0cm}<{\centering}|p{1.8cm}<{\centering}}
			\hline
			\hline
			Author & Algorithm & Modality & Accuracy (\%) \\
			\hline
			Depierre~\cite{depierre2018jacquard} & Jacquard & RGB-D & 74.2 \\
			\hline
			\multirow{2}*{Morrison~\cite{Morrison2020LearningRR}}& GG-CNN & D & 78 \\
		      & GG-CNN2 & D & 84 \\
			\hline
                \multirow{2}*{Teng~\cite{TENG2021107318}} & Depthwise & \multirow{2}*{D} & \multirow{2}*{88} \\
                & Conv. & & \\
                \hline
			\multirow{2}*{Kumra~\cite{kumra2020antipodal}} & \multirow{2}*{GR-ConvNet} & D & 93.7 \\
			& & RGB-D & 94.6 \\
	        \hline
            {Kumra~\cite{GR-ConvNetv2}} & {GR-ConvNetv2} & {RGB-D} & {\textbf{95.1}}\\
            \hline
			\multirow{2}*{Ours} & \multirow{2}*{PEGG-Net} & D & \textbf{93.8} \\
			& & RGB-D & 92.7 \\
			\hline
			\hline
		\end{tabular}
	\end{center}
\vspace{-4mm}
\end{table}

\subsection{Test Objects for Closed-loop Grasping in the Real-world}
\label{sec:VD}

We evaluate the performance of our network using three sets of objects: objects with adversarial geometry, household objects, and workshop tools. Fig. \ref{Objects} shows all the objects and the different backgrounds used during the experiments. We conduct three tests for each set of objects: a static grasping test, a dynamic grasping test, and a cluttered grasping test. During the static grasping test. Each object was tested in isolation for 10 different positions, orientations, and poses. For the dynamic grasping test, each object was tested in isolation and moved randomly during each of the 10 grasp attempts. Following the technique used by Morrison \textit{et al.}~\cite{morrison2018closing} to assist reproducibility, we translate each object by at least 10cm and rotate it by at least $25^\circ$. For the cluttered grasping test, we place all the objects of a set into the workspace and count the number of grasp attempts needed to clear all objects from the workspace. After each test is complete, we divide the number of objects in the test set by the number of successful grasp attempts needed to clear all objects from the workspace to obtain the grasp success rate. We consider a grasp attempt successful if the object can be successfully lifted to the dropping position outside the workspace.

\subsubsection{Objects With Adversarial Geometry}
This set consists of 13 3D-printed objects with adversarial geometry from the DexNet 2.0 dataset. All objects in this set have complex geometrical features. Even slight errors in the grasp prediction will result in a grasp failure. Therefore, this set of objects is very suitable for verifying the accuracy of the grasp poses generated by PEGG-Net. To the best of our knowledge, besides Mahler \textit{et al.}~\cite{mahler2017dexnet}, we are the only ones to evaluate our network on all 13 adversarial objects in the DexNet 2.0 dataset for 2D planar grasping.

\subsubsection{Household Objects}
This set consists of 14 everyday household items. To minimize redundancy, and to fairly evaluate the generalizability of novel objects, none of the objects in this set have similar shapes or sizes.

\subsubsection{Workshop Tools}
We select a set of 10 workshop tools to challenge our network. The depth difference between the workshop tools and the background is not as obvious as compared to the other sets of objects used in our experiments. In this context, we propose to conduct an experiment that previous works have not done: test the robustness of PEGG-Net in a scenario whereby the depth channel cannot provide clear geometric features for network prediction.


\subsection{Real-world Experiment Results}
\label{sec:VE}

Table \ref{Closed-loop} compares our real-world experiment results to previous works under static, dynamic, and cluttered conditions. Note that household objects are a vast category of objects, and prior works have each used different household objects for evaluating their networks. Therefore, the comparative performance for household objects listed in Table \ref{Closed-loop} is only indicative. PEGG-Net performs on par with GQ-CNN (which was trained on the adversarial objects) though ours is not being trained on the adversarial objects. PEGG-Net has also performed well when grasping the workshop tools with complex geometry, which further demonstrates its robustness. PEGG-Net can achieve better real-world grasping performance than the vast majority of other competitive networks, most of them much larger than PEGG-Net. We compute the overall grasp success rate by dividing the total number of successful grasp attempts by the total number of grasp attempts over all the grasp experiments we conducted. Overall, we have achieved a remarkable grasp success rate of 90.2\%, and PEGG-Net has achieved both a qualitative and quantitative breakthrough compared to previous works.

\begin{table}[t]
	\captionsetup{justification=centering}
	\renewcommand\arraystretch{1.0}
	\begin{center}
		\caption{Success rate of closed-loop grasping in various real-world scenes. All grasping experiments were conducted using a PEGG-Net model trained on the Jacquard dataset.}
		\label{Closed-loop}
		\begin{tabular}{p{0.9cm}<{\centering}|p{1.5cm}<{\centering}|p{1.3cm}<{\centering}|p{1.3cm}<{\centering}|p{1.3cm}<{\centering}}
			\hline
			\hline
			\multirow{2}*{Scenes} & \multirow{2}*{Methods} & Household objects & Adversarial Objects & Workshop tools \\
			& & (\%) & (\%) & (\%) \\
			\hline
			\multirow{6}*{Static} & Morrison~\cite{morrison2018closing} & 92.0 & 84.0 & -  \\
			 & Kumra~\cite{kumra2020antipodal} & 95.4 & - & - \\
			 & Lenz~\cite{lenz2014deep} & 89 & - & - \\
			 & Mahler~\cite{mahler2017dexnet} & 80.0 & 93.0 & - \\
			 & Pinto~\cite{pinto2015supersizing} & 73.0 & - & - \\
			 & Viereck~\cite{viereck2017learning} & \textbf{98.0} & - & - \\
              &  Xu ~\cite{GKNet} & 95.0 & 95.0 & - \\
			 & \multirow{2}*{Ours} & \underline{96.4} & \textbf{93.1} & \textbf{87.0} \\
			 & & \underline{(135/140)} & \textbf{(121/130)} & \textbf{(87/100)} \\
			 \hline
			\multirow{3}*{Dynamic} & Morrison~\cite{morrison2018closing}  & 88.0 & - & - \\
			& {Xu ~\cite{GKNet}} & {\textbf{96.7}} & {-} & {-} \\ 
			 & \multirow{2}*{Ours} & \underline{92.8} & \textbf{90.0} & \textbf{85.0} \\ 
			 & & \underline{(130/140)} & \textbf{(117/130)} & \textbf{(85/100)} \\
			\hline
			\multirow{3}*{Clutter} & Morrison~\cite{morrison2018closing}  & 87.0 & - & -  \\
			 & Viereck~\cite{viereck2017learning} & 89.0 & - & - \\
			 & {\multirow{2}*{Ours}} & {\textbf{89.7}} & {\textbf{86.1}} & {\textbf{90.1}} \\
			 & & {\textbf{(140/156)}} & {\textbf{(130/151)}} & {\textbf{(100/111)}} \\
			\hline
			\hline
		\end{tabular}
	\end{center}
\vspace{-6mm}
\end{table}

\section{Conclusions}
In this article, we present PEGG-Net, a novel real-time, pixel-wise grasp detection model that is robust in complex real-world scenes. We also design and implement a closed-loop control system that couples PEGG-Net for grasping in the real world. In the experimental section, we evaluate PEGG-Net on the Cornell and Jacquard datasets and achieved state-of-the-art performance. As a result, PEGG-Net has performed exceptionally well in our real-world grasping experiments using novel objects under static, dynamic and cluttered conditions. Both the benchmark results on publicly available grasping datasets and the real-world grasping experiments demonstrate our approach's detection accuracy, grasping robustness and efficiency.

\section{ACKNOWLEDGMENT}
This research is supported by the Agency for Science, Technology and Research (A*STAR) under its AME Programmatic Funding Scheme (Project \#A18A2b0046).

\bibliographystyle{IEEEtran}
\bibliography{root}

\end{document}